\documentclass[10pt,twocolumn,letterpaper]{article}

\usepackage[pagenumbers]{cvpr} % To force page numbers, e.g. for an arXiv version

\usepackage{graphicx}
\usepackage{tabularx}
\usepackage{epsfig}
\usepackage{epstopdf}
\usepackage{amsmath}
\usepackage{dsfont}
\usepackage{amssymb}
\usepackage{mathrsfs}
\usepackage{booktabs}
\usepackage{xcolor}
\usepackage{pifont}% http://ctan.org/pkg/pifont
\newcommand{\cmark}{\ding{51}}%
\newcommand{\xmark}{\ding{55}}%
\usepackage{multirow}
\usepackage[normalem]{ulem}
\graphicspath{{figures/}}
\usepackage{url}
\usepackage{wrapfig}
\usepackage{lipsum}
\usepackage{capt-of}% or \usepackage{caption}

\newcounter{example}%[section]
\renewcommand{\theexample}{\arabic{example}}
\newenvironment{example}{
        \vspace{1ex}
        \refstepcounter{example}
        {\noindent \bf [Example \theexample]}}{
        \eop\vspace{1ex}}
\newcommand{\eop}{\hspace*{\fill}\mbox{$\Box$}}

\usepackage[pagebackref,breaklinks,colorlinks]{hyperref}

\usepackage[capitalize]{cleveref}
\crefname{section}{Sec.}{Secs.}
\Crefname{section}{Section}{Sections}
\Crefname{table}{Table}{Tables}
\crefname{table}{Tab.}{Tabs.}

\def\cvprPaperID{5470} % *** Enter the CVPR Paper ID here
\def\confName{CVPR}
\def\confYear{2022}

\begin{document}

\title{MGA-VQA: Multi-Granularity Alignment for Visual Question Answering}

\author{\textbf{Peixi Xiong$^{1}$}\thanks{Work done during an internship at Samsung Research America.} \textbf{, Yilin Shen$^{2}$, Hongxia Jin$^{2}$}\\
${}^{1}$Northwestern University, ${}^{2}$Samsung Research America
}

\maketitle

%%%%%%%%% ABSTRACT
\begin{abstract}
Learning to answer visual questions is a challenging task since the multi-modal inputs are within two feature spaces. 
Moreover, reasoning in visual question answering requires the model to understand both image and question, and align them in the same space, rather than simply memorize statistics about the question-answer pairs.
Thus, it is essential to find component connections between different modalities and within each modality to achieve better attention.
Previous works learned attention weights directly on the features. 
However, the improvement is limited since these two modality features are in two domains: image features are highly diverse, lacking structure and grammatical rules as language, and natural language features have a higher probability of missing detailed information.
To better learn the attention between visual and text, we focus on how to construct input stratification and embed structural information to improve the alignment between different level components.
We propose Multi-Granularity Alignment architecture for Visual Question Answering task (MGA-VQA), which learns intra- and inter-modality correlations by multi-granularity alignment, and outputs the final result by the decision fusion module.
In contrast to previous works, our model splits alignment into different levels to achieve learning better correlations without needing additional data and annotations.
The experiments on the VQA-v2 and GQA datasets demonstrate that our model significantly outperforms non-pretrained state-of-the-art methods on both datasets without extra pretraining data and annotations. 
Moreover, it even achieves better results over the pre-trained methods on GQA.
\end{abstract}
%%%%%%%%%%%%%%%%%%%%%%%%%%%%%%%%%%%%
\begin{figure}[t]
\centering
\includegraphics[width=\columnwidth]{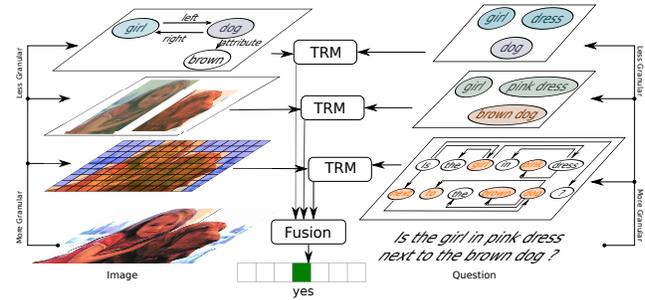}
 \caption{Multi-granularity alignment in Visual Question Answering. For each modality, the input is represented by different granularity levels, and Transformers (TRMs) are responsible for learning the correlations between levels.}
\label{fig:figure1}
\end{figure}
%%%%%%%%%%%%%%%%%%%%%%%%%%%%%%%%%%%%

\section{Introduction} 
\label{section-intro}
Visual Question Answering (VQA) continues to be a topic of interest due to its broad range of applications, such as the early education system and visual chatbot.
Multidisciplinary research on VQA has yielded innovations in multi-modality alignment~\cite{ilievski2017multi-modal}, natural language understanding~\cite{yi2018neural}, image understanding~\cite{goyal2017making}, and even multi-modal reasoning~\cite{cadene2019murel}.

Early methods addressed the multi-modal problem by the simple concatenation of visual features obtained from Convolutional Neural Network (CNN) and natural language features obtained from Recurrent Neural Network (RNN)~\cite{VQAV2}.
Although they established milestones in this field and have provided insights into merging features, such simple fusion methods do not offer good performance. 
This has motivated many additional related works that primarily focus on furtherly processing the features before merging, i.e., embedding these features into a new space~\cite{ren2015exploring} or utilizing an attention mechanism to achieve better alignment~\cite{nam2017dual}.
Another related research direction has involved the construction of graphs to represent image information~\cite{hudson2019learning}.
These explorations aimed to determine an effective way to achieve better alignment between multi-modal features;
however, the direct fusion of whole fine-grained image features and whole natural language sentences is complicated and lacks interpretability.

Taking Figure~\ref{fig:figure1} as an example, while a human can easily answer the question, a model cannot directly achieve that if without further processing the raw visual and language features for a better alignment.
However, if the multimodel components are correspondingly well-aligned, the model can work more straightforwardly to infer the answer.
More specifically, a machine that attends to specific words or concepts in the questions and specific visual components in the image would arguably be more robust to linguistic variations, irrelevant to the question's meaning and answer.

We tackle this problem by implementing granularity level alignment.
Apart from only determining ``where to look'' for visual and textual attention, we try to achieve ``looking correlation at the same level''.
One solution for achieving efficient attention-based alignment is to implement a Transformer (TRM).
Such architecture was first proposed as a machine translation model~\cite{vaswani2017attention}, and later popularly implemented in multi-modality-based tasks due to its good alignment ability.
However, many related works are pre-trained models, which require extra computational resources and extra human labor to collect additional data and its annotations (i.e., 3.3M images and captions for Conceptual Captions~\cite{sharma2018conceptual}). To make it even worse, for some domains, i.e., medical, it is hard to obtain and annotate such a large-scale external dataset.
% To achieve alignment more effectively, different from conventional Transformers, we embed graph-structured information into our model, involving lead graphs for extracting multi-modality features.
% Thus, unlike conventional Transformers, \ys{our focus is ...} we embed graph-structured information into our model, involving lead graphs for extracting multi-modality features to achieve alignment more effectively without extra data.
Thus, unlike conventional Transformers, our focus is to more effectively learn multi-modality alignment without extra data.
To achieve this, we embed graph-structured information into our model by involving the concept of lead graphs.\\

\textbf{Our main contributions are as follows:}
\begin{itemize}
% 	\item We propose a novel multi-granularity alignment architecture that jointly learns the intra- and inter-modality correlation and achieves interpretability.
	\item We propose a novel multi-granularity alignment architecture that jointly learns intra- and inter-modality correlations at three different levels: concept-entity level, region-noun phrase level, and spatial-sentence level. Moreover, the results are then integrated with the decision fusion module for the final answer.
	\item We propose a co-attention mechanism that jointly performs question-guided visual attention and image-guided question attention and improves interpretability.
% 	\item To better fuse the interaction between these two modalities, we proposed a decision fusion module.
	\item The experiments are conducted on two challenging benchmark datasets, GQA and VQA-v2.
	They demonstrate our model's effectiveness over the methods without extra pre-training data on both datasets.
	Moreover, our method even achieves better results over the pre-trained methods on GQA.
% 	validate the roles of different components in our model.
\end{itemize}

%%%%%%%%%%%%%%%%%%%%%%%%%%%
\begin{figure*}[tb!] %tb!
\centerline{\includegraphics[width=2.05\columnwidth]{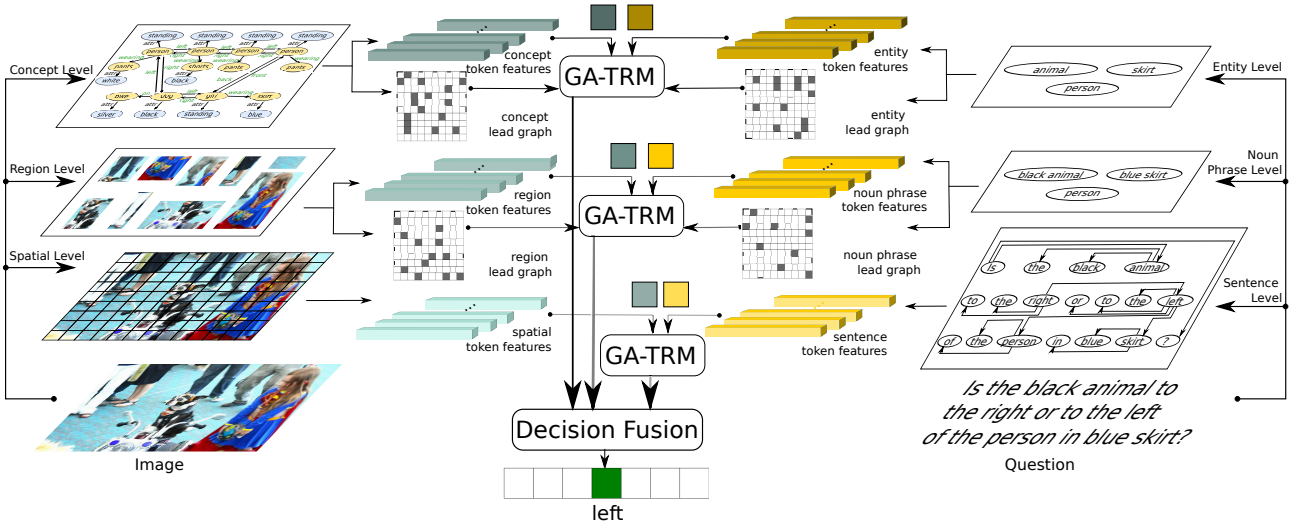}}
\caption{Overview of our approach. 
The following three granularity levels represent the image information: concept level, region level, and spatial level.
The input questions are described by entity level, noun phrase level, and sentence level information.
Later, token features and lead graphs are extracted from each level, then input into three granularity alignment Transformers (GA-TRMs) accordingly. 
(The lead graph in the figure is for demonstration, where its value is randomly set.)
Finally, the outputs are combined by the decision fusion module to obtain the final prediction.
} 
\label{fig-overview}
\end{figure*}
%%%%%%%%%%%%%%%%%%%%%%%%%%%
\section{Related works} 
\label{section-related}

\subsection{Visual Question Answering}
Visual Question Answering (VQA) is a popular joint visual language task that has been of increasing interest over the last few years.
Its main task is to answer questions about a provided image.
Similar to other multi-modal tasks, VQA requires the model to understand both the image and natural language and to combine their features of the two modalities.
However, there is a greater reliance on reasoning when answering visual questions. 
Implementing this reasoning ability is challenging, since the features of the two modalities lie in two very different domains: images lack the grammatical structure of the natural language, and it is high possible that the latter is biased in the sense that it makes it difficult for the model to output correct answers.
These further motivate the construction of a model that can align these two features properly.

Early solutions employed convolutional neural networks (CNNs) and recurrent neural networks (RNNs) to embed the image and question, respectively, and extracted features are them directly fused (i.e., via concatenation or a shallow network) to obtain the result~\cite{fukui2016multi-modal, zhou2015simple}.
In other works, one modality is further processed before fusion~\cite{xiong2016dynamic}, or both are jointly embedded into a new space by additional networks~\cite{ren2015exploring}.
Yet other works proposed architectures that imply element-wise summation~\cite{lu2016hierarchical, yang2016stacked} or multiplication~\cite{li2016visual,nam2017dual} to achieve better fusion of multi-modal features.
The simple feature fusion methods established milestones in VQA task, providing insights into merging multi-modal features.

% To better evaluate model performance, many related VQA datasets have been published~\cite{malinowski2014multi, VQAV2, kafle2017analysis}. 
% In addition to general datasets, many others focus on evaluating performance regarding specific aspects of the problem.
% i.e., GQA~\cite{hudson2019gqa}, which contains logical questions that require the ability to reason and understand scenes, and VQA-CP~\cite{agrawal2018don}, which targets language biases.
% Another dataset, Visual Genome~\cite{krishna2017visual}, focuses mainly on structured image concepts.

\subsection{Co-attention mechanisms in VQA}
Many works focus on exploring image attention models for VQA~\cite{zhu2016visual7w, yang2016stacked, andreas2015deep}. 
In the natural language processing domain, there are many related works on modeling language attention~\cite{yin2016abcnn, bahdanau2014neural, rocktaschel2015reasoning}.
Some works learn textual attention for questions and visual attention for images simultaneously.
% ~\cite{yu2018beyond} describes a bilinear pooling method that achieves effective fusion of multi-modal features by exploiting their correlations sufficiently.
~\cite{lu2016hierarchical} presents a hierarchical co-attention model for VQA that jointly reasons image and question attention.
~\cite{nam2017dual} is a multi-stage co-attention learning model that refines the attention based on the memory of the previous attention.
% ~\cite{gao2019dynamic} proposed a novel model that dynamically fuses multi-modal features with intra- and inter-modality information flow, which alternatively passes dynamic information between and across the visual and language modalities.
~\cite{yu2019deep} proposed a deep Modular Co-Attention Network that consists of Modular Co-Attention layers cascaded in depth.
These works focus on the alignment between text features and fine-grained image features, where images lack language structure and grammatical rules, leading to difficulty in obtaining a good result.
In addition, most of these works process questions in a simple manner and ignore the inner logical relations in the natural language domain.
These issues become a bottleneck for understanding the relationships between multi-modal features. 

\subsection{Transformers in the multi-modal task}
The Transformer~\cite{vaswani2017attention} was initially proposed as a powerful machine translation model.
However it gained substantial attention due to its ability to learn attention in all positions.
In contrast to recurrent neural networks, Transformer expands the ability of the model to focus on the inner relations between sentences,producing a so-called ``self-attention'' property. (i.e., demonstrative term in the sentence)  
For example, when translating the sentence “The animal didn’t cross the street because it was too tired,” it would be helpful to know which word ``it'' refers to, as this would greatly improve the translation result.
Due to the self-attention property, the Transformer architecture has been applied to other tasks beyond image captioning~\cite{lu2019vilbert} or visual question answering~\cite{tan2019lxmert}; it also shows up in the works of vision-and-language navigation~\cite{hao2020towards} and video understanding~\cite{sun2019videobert}.
Furthermore, implementation of the Transformer architecture aids the model in learning cross-modal representations from a concatenated sequence of visual region features and language token embeddings~\cite{su2019vl,li2020unicoder}.
In addition, it learns joint representations that are appropriately contextualized in both modalities.
However, these alignments are always achieved by pre-training on additional dataset~\cite{sharma2018conceptual,lin2014microsoft, vicente2016large}.

\section{Approach} 
\label{section-approach}
We now introduce our Multi-Granularity Alignment Transformer for VQA (MGA-VQA).
Our main idea of the model is to align multiple information levels accordingly between multi-modal inputs and to integrate the information to obtain the final prediction.
\figurename~\ref{fig-overview} illustrates the architecture of our proposed model, which consists of three sets of alignments with different granularity levels.
First, objects are detected from the input image, with their names, corresponding attributes, and relations. 
On the question side, noun phrases, entities, and sentence grammatical structure are detected.
Then lead graphs are used to furtherly guide alignment learning, and they are constructed from the structural information extracted in the above steps, where the nodes in the graphs are regarded as the token features for the next steps. 
These features are the basic components of the following three levels of granularity alignment transformers (GA-TRMs): information of the concept level and entity level, information of the region level and noun phrase level, and information of the spatial level and sentence level.
% The lead graphs are then used to assist in co-attention learning.
Finally, the outputs of the three GA-TRMs are used to predict the answer via the decision fusion module.

Section~\ref{section-features} describes the extraction of the different granularity level features, construction of the lead graphs, and formation of the token features from the image and question.
Section~\ref{section-trm} explains in detail how the GA-TRMs use the token features and lead graphs.
Section~\ref{section-fusion} illustrates the merging of the outputs from the three GA-TRMs by the fusion module.
% Section~\ref{section-details} shows the implementation details of our model.
\subsection{Granularity levels in VQA} 
\label{section-features}
In our model, three granularity levels are set for image and question, describing different levels of information.
They are used to be aligned to the corresponding levels between modalities, i.e., concept-entity, region-noun phrase, and spatial-sentence.
\subsubsection{Granularity information in image}
Given an input image ($Img$), three levels of features are extracted with different levels of granularity.
For each granularity level, there are an associated set of token features and lead graph pairs.
We first construct graph $\mathcal {G} = \{E, L\}$ representing the current granularity level information, where the tokens ($E$) are the entities in the graph, and the lead graph ($L$) is the connection pairs in the corresponding adjacency matrix.
We now describe each level in detail as follows.\\

\noindent \textbf{Concept Level}\\
The concept level consists of the semantic features of objects, attributes, and relations between objects.
We first extract this information from the image and build the corresponding graph $\mathcal {G}_c$, i.e., the top left section in Figure~\ref{fig:figure1}.
To better input $\mathcal {G}_c$ to the next stage, we first regard relations as extra nodes; then, we split this graph into node sequence ($E_c$) and pairs that represent node connections by index ($L_c$).

% The directional connections are consist of index pairs, describing both the ``subject-predicate-object'' relation and the ``subject/object-attribute'' information.
% Additionally, to describe the relation, We split the ``subject-predicate-object'' triple into ``subject $\rightarrow$ predicate'' and ``predicate $\rightarrow$ object'' pair so that it is easy to use index pair.

To describe the relations between nodes, we split the ``subject-predicate-object'' triple into ``subject $\rightarrow$ predicate'' and ``predicate $\rightarrow$ object'' pair. 
In this way, both the ``subject-predicate-object'' relation and the ``subject/object-attribute'' information are described by the index pairs, which are regarded as the node connections ($L_c$).

\begin{example}
\label{exm-1}
For Figure~\ref{fig:figure1}, there are\\
$E_c = [girl, left, right, dog, brown]$, \\
$L_c = [(0,1), (1,3), (3, 2), (2, 1), (3, 4)]$.
\end{example}

\noindent Note that the token sequence feature $\textbf{T}_c = \{\textbf{t}_{c_1}, \textbf{t}_{c_2}, ..., \textbf{t}_{c_N}\}$ is computed from node sequence $E_c = \{e_{c_i}\}_{i=1}^{N}$ by GloVe embedding~\cite{Pennington14glove:global} and the Multi-layer perceptron (MLP). \\

\noindent \textbf{Region Level}\\
The region level describes the middle-level visual features, which represents the visual region of object.
Unlike the object features in the concept level, features in this level describe the object information visually instead of semantically.
The token sequence features $\textbf{T}_r = \{\textbf{t}_{r_i}\}_{i=1}^{M}$ are extracted by the Faster R-CNN method, and the relation pairs $L_r$ are similar to $L_c$, where if there is a semantic relation between two objects at the concept level, there is a corresponding relation pair at the region level.

\begin{example}
\label{exm-2}
For Figure~\ref{fig:figure1}, there are $\textbf{T}_r = [\textbf{t}_{girl}, \textbf{t}_{dog}]$ and $L_r = [(0,1), (1,0)]$.
\end{example}

% for the previous example, $\textbf{T}_r = [\textbf{t}_{girl},\textbf{t}_{dog}]$ and $L_r = [(0,1), (1,0)]$.\\

\noindent \textbf{Spatial Level}\\
The spatial level describes the holistic but highest granularity visual features and provides detailed, spatial and supplementary information to the previous two levels, i.e., scene information.
% Such fine-grained features capture pixel-wise information, complementing details missing from the other two levels of features and involving spatial information.
The token sequence features $\textbf{T}_{sp}$ are extracted from the backbone CNN, and $L_{sp}$ is equal to the fully connected relations for all feature cells.

\subsubsection{Granularity information in question}
Similar to $Img$, three levels of granularity are extracted from input question ($Q$).
Most previous works only focused on the image features, ignoring inter-correlation within sentence.
In contrast to them, our proposed method extracts structural and grammatical information, for better alignments.\\

\noindent \textbf{Entity Level}\\
The entity-level features represent individual objects in $Q$ without attributes and help the model to achieve alignment in the abstract.
The token features $\textbf{T}_e$ are processed in a similar manner as concept features for $Img$, and the corresponding lead graph pair $L_e$ corresponds to the fully connected pair.\\

\noindent \textbf{Noun Phrase Level}\\
We filter the result from a constituency parser for our noun phrase level, discarding the determiners (e.g., 'a', 'the') and filtering out the words expressing positional relations (e.g., 'left', 'right') to save computational resources.
The noun phrase level is constructed to align with the region-level features in $Img$, where the visual features contain attributions.
Since most of the components are composed of multiple words, instead of merging them into a single token, we split them and process the GloVe features with the MLP, and obtain the token features as $\textbf{T}_{np}$. 
In addition, the corresponding lead graph pair $L_{np}$ corresponds to the fully connected pair.\\

\noindent \textbf{Sentence Level}\\
For sentence level, we process $Q$ with the dependency parser and get the corresponding adjacency matrix ($Dep_{s}$) from the dependency graph.
Since visual features are of a higher level and require less context aggregation than natural language~\cite{NEURIPS2019_c74d97b0}, sentence-level features need to be furtherly processed before alignment to better embed the structural information with the input.
Instead of directly input the sentence token into the Transformer to fuse multi-modality features, we first use an extra Transformer module to process the sentence to get the context-aware features $\textbf{T}_{s}$.
\begin{equation}
\textbf{T}_{s} = \mathcal Trm(\mathcal {MLP}(\mathcal GloVe(Q)), Dep_{s}),
\end{equation}
% The token features are $T_{s} = \mathcal Trm(\mathcal {MLP}(\mathcal GloVe(Q)), Dep_{s})$, 
where $\mathcal GloVe(\cdot)$ is the GloVe word embedding, $\mathcal {MLP}(\cdot)$ is the Multi-layer perceptron, and $\mathcal Trm(t, g)$ is the Transformer module with input tokens $t$ and attention mask $m$.
Since the connection information is already embedded in $\textbf{T}_{s}$, the lead graph pair for sentence level ($L_{s}$) consists of the fully connected pair.
Details are shown in the Section~\ref{section-details}.

\subsection{Multi-modality granularity alignment} 
\label{section-trm}
%%%%%%%%%%%%%%%%%%%%%%%%%%%%%%%%%%%%
\begin{figure}[t]
\centering
\includegraphics[width=0.6\columnwidth]{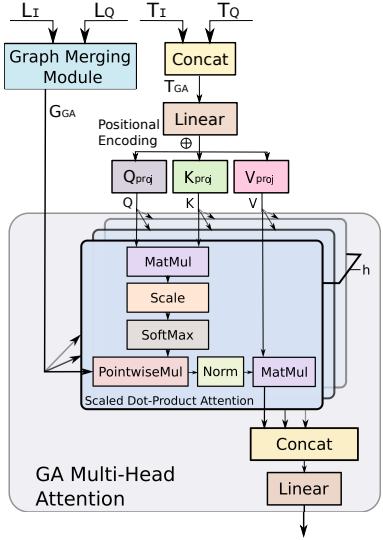}
 \caption{Granularity Alignment (GA) Attention.}
\label{fig:trm}
\end{figure}
%%%%%%%%%%%%%%%%%%%%%%%%%%%%%%%%%%%%
In this section, we explain our design for using token features and lead graphs for alignment learning.
For the lowest granularity level, the concept and entity token features from $Img$ and $Q$ respectively, are used as the token inputs of the GA-TRM, which obtains the most abstract information of both modalities.
For the middle granularity level, the object region and noun phrase token features from $Img$ and $Q$, respectiively, are fed into the GA-TRM to learn the co-attention.
For the highest granularity level, the spatial token features from $Img$ and the sentence token features from $Q$ are aligned, supporting the model with the most detailed information from the two modalities.
This level of alignment is responsible for learning the cross attention between the detailed information, which does not show in the previous two levels of features.

\subsubsection{GA Multi-head Attention}
The Transformer architecture was initially proposed in~\cite{NIPS2017_3f5ee243}, in which it used stacked self-attention and point-wise, fully connected layers for both the encoder and decoder.
The attention function can be described as mapping a query and a set of key-value pairs to an output.
Our model follows the same architecture, as shown in detail in Figure~\ref{fig:trm}.
The token features from the image ($\textbf{T}_I \in \{\textbf{T}_c, \textbf{T}_r, \textbf{T}_{sp}\}$) and question ($\textbf{T}_Q \in \{\textbf{T}_e, \textbf{T}_{np}, \textbf{T}_s\}$) modalities are concatenated.
After linear projection, they then employed with the learnable positional encoding~\cite{NIPS2017_3f5ee243} to include both relative and absolute position information.
For each token, a query vector ($\textbf{Q}$), a key vector ($\textbf{K}$), and a value vector ($\textbf{V}$) are created, by multiplying the embeddings of the three matrices that are trained during the training process.
Instead of utilizing a single attention module, we also linearly project $\textbf{Q}$, $\textbf{K}$, and $\textbf{V}$  $h$ times with different, learned linear projections.
Each of the sets of vectors is then input into the scaled dot-product attention, and point-wise multiplied with the lead graph ($\textbf{G}_{GA}$) from the graph merging module:
\begin{equation}
\text{Att}_{GA}(\textbf{Q}, \textbf{K}, \textbf{V} ) = norm(\text{softmax}(\frac{\textbf{Q}\textbf{K}^T}{\sqrt{ d_k }})
\circ \textbf{G}_{GA}) \textbf{V},
\end{equation}
where $d_k$ represents the dimensionality of the input, and $norm(\cdot)$ is the normalization over rows.
Then, they are concatenated and once again projected, resulting in the final values.

\subsubsection{Graph Merging Module}
Our graph merging module is designed to convert graph pairs ($L$) to single modality lead graphs ($ \{\textbf{G}_{I}, \textbf{G}_{Q}\}$), and then merge them into the multi-modality lead graph ($\textbf{G}_{GA}$).\\

\noindent \textbf{Single modality lead graphs generation}\\
The single modality lead graphs from image ($\textbf{G}_I$) and question ($\textbf{G}_Q$) are binary graphs that are first constructed from the corresponding graph pairs from images ($L_I \in \{L_c, L_r, L_{sp}\}$) and questions ($L_Q \in \{L_e, L_{np}, L_s\}$).
% Here $L_I$ and $L_Q$ are chosen from the lead graph pairs from images ($L_c$, $L_r$, $L_{sp}$) and questions ($L_e$, $L_{np}$, $L_s$), correspondingly.
The dimension of the $\textbf{G}_I$ is $||\textbf{T}_I|| \times
 ||\textbf{T}_I||$, while that of $\textbf{G}_Q$ is  $||\textbf{T}_Q|| \times
 ||\textbf{T}_Q||$.
For each pair in $L_I$ and $L_Q$, we assign the corresponding cell in the binary graph a value of 1, while the others are assigned a value of 0.

\begin{example}
\label{exm-2}
If $L_I = [(0,1), (1,3), (3, 2), (2, 1)]$ and $||\textbf{T}_I|| = 4$, 
then there is: \\
% $\textbf{G}_I = [[0,1,0,0],[0,0,0,1],[0,1,0,0],[0,0,1,0]]$, 
$ \displaystyle
    \begin{aligned} 
       \hspace{16ex} \textbf{G}_I &= \begin{bmatrix}0 & 1 & 0 & 0 \\\ 0 & 0 & 0 & 1 \\\ 0 & 1 & 0 & 0 \\\ 0 & 0 & 1 & 0\end{bmatrix}
    \end{aligned}
  $ \\
where the dimension of $\textbf{G}_I$ corresponds to $||\textbf{T}_I||$, and the 2-D index of the nonzero value in $\textbf{G}_I$ corresponds to $L_I$.
\end{example}

% e.g., $L_I = [(0,1), (1,3), (3, 2), (2, 1)]$, and $||\textbf{T}_I|| = 4$, then $\textbf{G}_I = [[0,1,0,0],[0,0,0,1],[0,1,0,0],[0,0,1,0]]$, where the dimension of $\textbf{G}_I$ corresponds to $||\textbf{T}_I||$, and the 2-D index of the nonzero value in $\textbf{G}_I$ corresponds to $L_I$.\\

\noindent \textbf{Multi-modality lead graphs generation}\\
The multi-modality lead graphs ($\textbf{G}_{GA}$s) are a set of binary graphs of dimension $(||\textbf{T}_I|| + ||\textbf{T}_Q||) \times (||\textbf{T}_I|| + ||\textbf{T}_Q||)$.
We set different lead graphs for different layers of encoders.

For the first layer, the lead graph is
\begin{equation}
\begin{bmatrix}\textbf{0}_{||\textbf{T}_I|| \times ||\textbf{T}_I||} & \textbf{0}_{||\textbf{T}_I|| \times ||\textbf{T}_Q||} \\\textbf{0}_{||\textbf{T}_Q|| \times ||\textbf{T}_I||} & \textbf{\textbf{1}}_{[||\textbf{T}_Q|| \times ||\textbf{T}_Q||]} \end{bmatrix}.
\end{equation}
\noindent This is to make the model learn the self-attention of the question, since the visual features are relatively high-level and require limited context aggregation with respect to words in a sentence, the latter of which needs further processing.

For the second layer, the lead graph is
% $[[\textbf{0}_{||\textbf{T}_I|| \times ||\textbf{T}_I||},\textbf{1}_{||\textbf{T}_I|| \times ||\textbf{T}_Q||}],[\textbf{1}_{||\textbf{T}_Q|| \times ||\textbf{T}_I||},\textbf{0}_{||\textbf{T}_Q|| \times ||\textbf{T}_Q||}]]$.
\begin{equation}
\begin{bmatrix}\textbf{0}_{||\textbf{T}_I|| \times ||\textbf{T}_I||} & \textbf{1}_{||\textbf{T}_I|| \times ||\textbf{T}_Q||} \\\textbf{1}_{||\textbf{T}_Q|| \times ||\textbf{T}_I||} & \textbf{\textbf{0}}_{[||\textbf{T}_Q|| \times ||\textbf{T}_Q||]} \end{bmatrix}.
\end{equation}
\noindent This is to make the model learn the co-attention between the modalities.

For the third layer, the lead graph is
% $[[\textbf{G}_{I},\textbf{1}_{||\textbf{T}_I|| \times ||\textbf{T}_Q||}],[\textbf{1}_{||\textbf{T}_Q|| \times ||\textbf{T}_I||},\textbf{G}_{Q}]]$, 
\begin{equation}
\begin{bmatrix}\textbf{G}_{I} & \textbf{1}_{||\textbf{T}_I|| \times ||\textbf{T}_Q||} \\\textbf{1}_{||\textbf{T}_Q|| \times ||\textbf{T}_I||} & \textbf{G}_{Q} \end{bmatrix}.
\end{equation}
which makes the encoder focus on the existing connectivity in the two modalities.\\

\subsection{Multi-granularity decision fusion} 
\label{section-fusion}
The outputs of each level alignment are $\textbf{H}_{ce}$, $\textbf{H}_{rn}$ and $\textbf{H}_{ss}$, which embed the alignments of concept-entity, region-noun phrase and spatial-sentence, respectively.
We define the linear multi-modal fusion function as follows:
% \begin{equation}
% \mathring{\textbf{H}}_{ce} = \textbf{W}^{T}_{ce}~\mathcal LayerNorm(\textbf{H}_{ce}),
% \end{equation}
% \begin{equation}
% \mathring{\textbf{H}}_{rn} = \textbf{W}^{T}_{rn}~\mathcal LayerNorm(\textbf{H}_{rn}),
% \end{equation}
% \begin{equation}
% \mathring{\textbf{H}}_{ss} = \textbf{W}^{T}_{ss}~\mathcal LayerNorm(\textbf{H}_{ss}),
% \end{equation}
\begin{equation}
\mathring{\textbf{H}}_{l} = \textbf{W}^{T}_{l}~\mathcal LayerNorm(\textbf{H}_{l}), l \in \{ce, rn, ss\}
\end{equation}
\begin{equation}
\textbf{H}_{GA} = \textbf{W}^{T}_{GA}~[\mathring{\textbf{H}}_{ce} ; \mathring{\textbf{H}}_{rn} ; \mathring{\textbf{H}}_{ss}],
\end{equation}
where $[\cdot;\cdot;\cdot]$ is the concatenation operation on vectors, $\textbf{W}_{ce}$, $\textbf{W}_{rn}$, $\textbf{W}_{ss}$ and $\textbf{W}_{GA}$ are linear projection matrices, and $\mathcal LayerNorm(\cdot)$ is used to stabilize the training~\cite{ba2016layer}.

In this work, we regard the VQA problem as a classification problem, and the final outputs are predicted by averaging the below logits.
For the loss, we individually compute the cross-entropy loss from the three alignment streams, and make each alignment stream equally contribute to the loss.
We define the loss as follows:
\begin{equation}
L = \mathcal L_{\text{CE}}(f_{ce}, a) + \mathcal L_{\text{CE}}(f_{rn}, a) + \mathcal L_{\text{CE}}(f_{ss}, a) + \mathcal L_{\text{CE}}(f_{GA}, a),
\label{eq:loss}
\end{equation}
where $a$ is the answer of the question.
And $f_{ce}$, $f_{rn}$, $f_{ss}$, and $f_{GA}$ represent the logits for the above three streams and their fusion, respectively, which are linearly projected by the previous outputs ($\textbf{W}_{ce}$, $\textbf{W}_{rn}$, $\textbf{W}_{ss}$ and $\textbf{W}_{GA}$).

\section{Experimental setup and results}
\label{section-exp}
%%%%%%%%%%%%%%%%%%%%%%%%%%%%%%%%%%%%%%%%%%
\begin{table*}[tbh]
\centering
% \resizebox{\textwidth}{!}{%
\begin{tabular}{@{}llc|llll|llll@{}}
\toprule
\multicolumn{3}{c}{\textbf{Method}}    & \multicolumn{8}{|c}{\textbf{Performance}} \\ 
\toprule
\textbf{No.} & \textbf{Name}    & \textbf{Extra Data} &  \textbf{Yes/No} & \textbf{Number} & \textbf{Other} & \textbf{Test-dev}  & \textbf{Yes/No} & \textbf{Number} & \textbf{Other} & \textbf{Test-std}   \\ \hline
10     & MUTAN~\cite{ben2017mutan}        & \xmark & 85.14 & 39.81 & 58.52 & 67.42 & - & - & - & 67.36    \\
11     & Counter~\cite{zhang2018learning} & \xmark & 83.14 & 51.62 & 58.97 & 68.09 & 83.56 & 51.39 & 59.11 & 68.41    \\
12     & BLOCK~\cite{ben2019block}        & \xmark & 83.60 & 47.33 & 58.51 & 67.58 & 83.98 & 46.77 & 58.79 & 67.92    \\
13     & MuRel~\cite{cadene2019murel}     & \xmark & 84.77 & 49.84 & 57.85 & 68.03 & - & - & - & 68.41    \\
14     & ViLBERT~\cite{lu2019vilbert}     & \cmark & - & - & - & 70.55 & - & - & - & 70.92 \\
15     & MCAN~\cite{yu2019deep}             & \xmark & 86.82 & 53.26 & 60.72 & 70.63 & - & - & - & 70.90    \\
16     & VisualBERT~\cite{li2019visualbert}       & \cmark & - & - & - & 70.80 & - & - & - & 71.00     \\ 
17     & VL-BERT~\cite{su2019vl}          & \xmark & - & - & - & 69.58 & - & - & - & 70.90     \\ 
18     & VL-BERT~\cite{su2019vl}          & \cmark & - & - & - & 71.79 & - & - & - & 72.22 \\   
19     & Oscar~\cite{li2020oscar}         & \cmark & - & - & - & 73.61 & - & - & - & 73.82 \\ \hline
20     & Ours~\footnotemark[1]            & \xmark & 86.84 & 54.36 & 60.97 & 70.99 & 87.13 & 55.10 & 61.20 & 71.28    \\ \bottomrule
\end{tabular}
% }
\caption{Overall accuracy on the VQA-v2 dataset.}
\label{table:vqa}
\end{table*}
%%%%%%%%%%%%%%%%%%%%%%%%%%%%%%%%%%%%%%%%%%
%%%%%%%%%%%%%%%%%%%%%%%%%%%%%%%%%%%%%%%%%%
\begin{table*}[tbh]
\centering
% \resizebox{\textwidth}{!}{%
% \begin{tabular}{lllllll|c}
\begin{tabular*}{\textwidth}{p{0.012\linewidth}p{0.1\linewidth}p{0.1\linewidth}p{0.1\linewidth}p{0.1\linewidth}p{0.1\linewidth}p{0.1\linewidth}p{0.1\linewidth}|c}
\toprule
\textbf{No.} & \textbf{Single Modality} & \textbf{Multi-Modality} & \textbf{Low Level GA} & \textbf{Middle Level GA} & \textbf{High Level GA} & \textbf{Lead Graph} & \textbf{Node Reduction} & \textbf{Accuracy} \\ \hline
21    & \xmark    & \cmark    & \cmark    & \xmark    & \xmark    &  \cmark    & \xmark & 48.01   \\
22    & \xmark    & \cmark    & \xmark    & \cmark    & \xmark    &  \cmark    & \xmark & 54.99   \\
23    & \xmark    & \cmark    & \xmark    & \xmark    & \cmark    &  \cmark    & \xmark & 61.65   \\
24    & \cmark    & \xmark    & -         & -         & -         &  \cmark    & \xmark & 50.22    \\
25    & \xmark    & \cmark    & \cmark    & \cmark    & \cmark    &  \xmark    & \xmark & 61.45   \\ 
26    & \xmark    & \cmark    & \cmark    & \cmark    & \cmark    &  \xmark    & \cmark & 62.78   \\ \hline
9     & \xmark    & \cmark    & \cmark    & \cmark    & \cmark    &  \cmark    & \xmark & 65.93   \\ \bottomrule
\end{tabular*}%
% }
\caption{Ablation study on the GQA dataset.}
\label{table:ablation}
\end{table*}
%%%%%%%%%%%%%%%%%%%%%%%%%%%%%%%%%%%%%%%%%%
%%%%%%%%%%%%%%%%%%%%%%%%%%%%%%%%%%%%%%%%%%
\begin{table}[tbh]
\centering
% \resizebox{0.8\textwidth}{!}{%
\begin{tabular*}{0.47\textwidth}{p{0.012\linewidth}p{0.21\linewidth}p{0.1\linewidth}|ccc} 
\toprule
\multicolumn{3}{c|}{ \textbf{Method} } & \multicolumn{3}{c}{\textbf{Performance} }    \\ 
\toprule
\textbf{No.}  & \textbf{Name}   & \textbf{Extra}     & \textbf{Open}  & \textbf{Binary}  & \textbf{Overall}   \\ 
\hline
1    & BUA~\cite{anderson2018bottom} & \xmark    & 34.83     & 66.64     & 49.74    \\
2    & LCGN~\cite{hu2019language} & \xmark    & -     & -     & 56.10    \\
3    & NSM~\cite{hudson2019learning} & \xmark    & 49.25     & 78.94     & 63.17    \\
4    & LXMERT~\cite{tan2019lxmert} & \cmark    & -     & -     & 60.33    \\
5    & LRTA~\cite{liang2020lrta}  & \xmark    & -     & -     & 54.48    \\
6    & Oscar~\cite{li2020oscar} & \cmark    & -     & -     & 61.62    \\
7    & VinVL~\cite{zhang2021vinvl} & \cmark    & -     & -     & 65.05    \\
8    & Human~\cite{hudson2019gqa} & \xmark    & 87.40     & 91.20     & 89.30    \\ 
\hline
9    & Ours & \xmark    &54.29     & 78.25     & 65.93    \\
\bottomrule
\end{tabular*}
% }
\caption{Overall accuracy on the GQA dataset.}
\label{table:gqa}
\end{table}
%%%%%%%%%%%%%%%%%%%%%%%%%%%%%%%%%%%%%%%%%%

\subsection{Datasets}
\subsubsection{VQA-v2}
% \noindent \textbf{VQA-v2}
The VQA~\cite{balanced_vqa_v2} dataset is a real image dataset that contains over 204k images from COCO, 614k free-form natural language questions, and over 6 million free-form answers. 
For each question, 10 answers were gathered for robust inter-human variability.
To be consistent with ‘human accuracy,’ the accuracy metric is $acc(ans) = min\{\frac{\#~humans~that~provided~that~answer}{3}, 1\}$, showing that an answer is regarded as 100\% accurate if at least three annotations exactly match the predicted answer.

\subsubsection{GQA}
% \noindent \textbf{GQA}
The GQA~\cite{hudson2019gqa} dataset consists of 20 million compositional questions involving a diverse set of reasoning skills and 1.5 million questions with closely controlled answer distributions. 
% In contrast to other synthetic datasets, GQA has a large space of possible objects and attributes, avoiding high-capacity models that memorize all combinations from easily performing their tasks and increasing effective compositionality. 
Compared with other real-image VQA datasets, the GQA dataset contains fewer language biases, involves more reasoning, and focuses on large vocabulary questions.
We evaluate the reasoning ability of our MGA-VQA with the test split and conduct ablation studies to validate the effectiveness of each module in our model.

\subsection{Implementation details}
\label{section-details}
Our model setting is based on~\cite{vaswani2017attention}.
The encoder and decoder are separately composed of a stack of 3 identical layers, while the decoder is also composed of a stack of 3 identical layers.
For the multi-head attention, we set 8 heads to achieve co-attention learning.
The model is trained with distributed training in PyTorch, with 4 GeForce RTX 3090 GPUs.
The learning rate is set to $10^{-4}$ with Adam optimizer, and batch size is set to 256.
We merge same relation tokens and attribute tokens in the concept level to reduce computational load and update the lead graph accordingly; however, we do not change object category tokens.
The [SEP] (special end-of-sequence) token is inserted after token features from image modality and is included in the corresponding dimension.
The visual features are extracted from BUA~\cite{anderson2018bottom}, and the scene graph is built in a similar manner as~\cite{kim2020hypergraph}. 
The spatial level features are obtained from Resnet-101 backbone~\cite{he2016deep}.

% \subsection{Pretraining and non-pretraining}
% Our model mainly focuses on non-pretraining since pre-training mechanism requires more computational resources (generally contain millions of images and captions) and extra supervised data.
% Moreover, there is no available data for the model to pre-train the alignment between visual features and question features in different granularities.
% Our model learns the multi-grained data without explicit pre-training with additional labels.
% In experiments, our model outperforms state-of-the-art methods on the GQA dataset and non-pretraining methods on the VQA-v2 dataset, and it achieves competitive results with pre-trained models that need extra datasets and computational resources.

% \ys{
\subsection{Model training details}
Our model is designed to only use visual question answer annotation pairs to learn finer grained latent features.
% However, existing pre-training model relies on using the additional labeled finer grained granularity (e.g., object) to explicitly learn finer grained features.
However, existing pre-training models rely on the additional labeled data describing finer-grained granularity (e.g., captioning) to explicitly learn finer-grained features, which require more computational resources and extra supervised data. 
%existing method relies on XX, and XX has these limitations/disadvantages. Thus, our experiment is focused on ...
Thus, our experiments are focused on training our model from scratch (a.k.a., non-pretraining).
Nevertheless, our method not only achieves significant performance gain over existing non-pretraining methods, but also achieves competitive performance with some of pretraining methods.
Next, we will show you detailed results.
% }

\subsection{Quantitative results on GQA}
\subsubsection{Overall performance}
We evaluate our MGA-VQA model on the GQA dataset with the top-1 accuracy, as shown in Table~\ref{table:gqa}. 
The methods that need the extra dataset~\cite{sharma2018conceptual,lin2014microsoft, vicente2016large} to pre-train the Transformer model are marked in the "Extra Data" column of the table.
Among the evaluated methods, BUA~\cite{anderson2018bottom} is an attention-based model that enables calculation of attentiion at the region level rather than by using a uniform grid of equally sized image regions.
LCGN~\cite{hu2019language}, NSM~\cite{hudson2019learning}, and LRTA~\cite{liang2020lrta} mainly focus on solving complicated visual questions by first constructing graphs that represent the underlying semantics.
LXMERT~\cite{tan2019lxmert}, Oscar~\cite{li2020oscar}, and VinVL~\cite{zhang2021vinvl} are Transformer-based models that solve the visual language problem by pre-training the model to align visual concepts and corresponding concepts in the text modality.
The table shows that our model outperforms the state-of-the-art methods, even those need pre-training.
\footnotetext[1]{For a fair comparison, we do not use validation split and other external VQA data (i.e., Visual Genome~\cite{krishna2017visual}) to train the model.}

%%%%%%%%%%%%%%%%%%%%%%%%%%%%%%%%%%%%%%%%%%
\begin{figure}[tb!] %tb!
\includegraphics[width=\columnwidth]{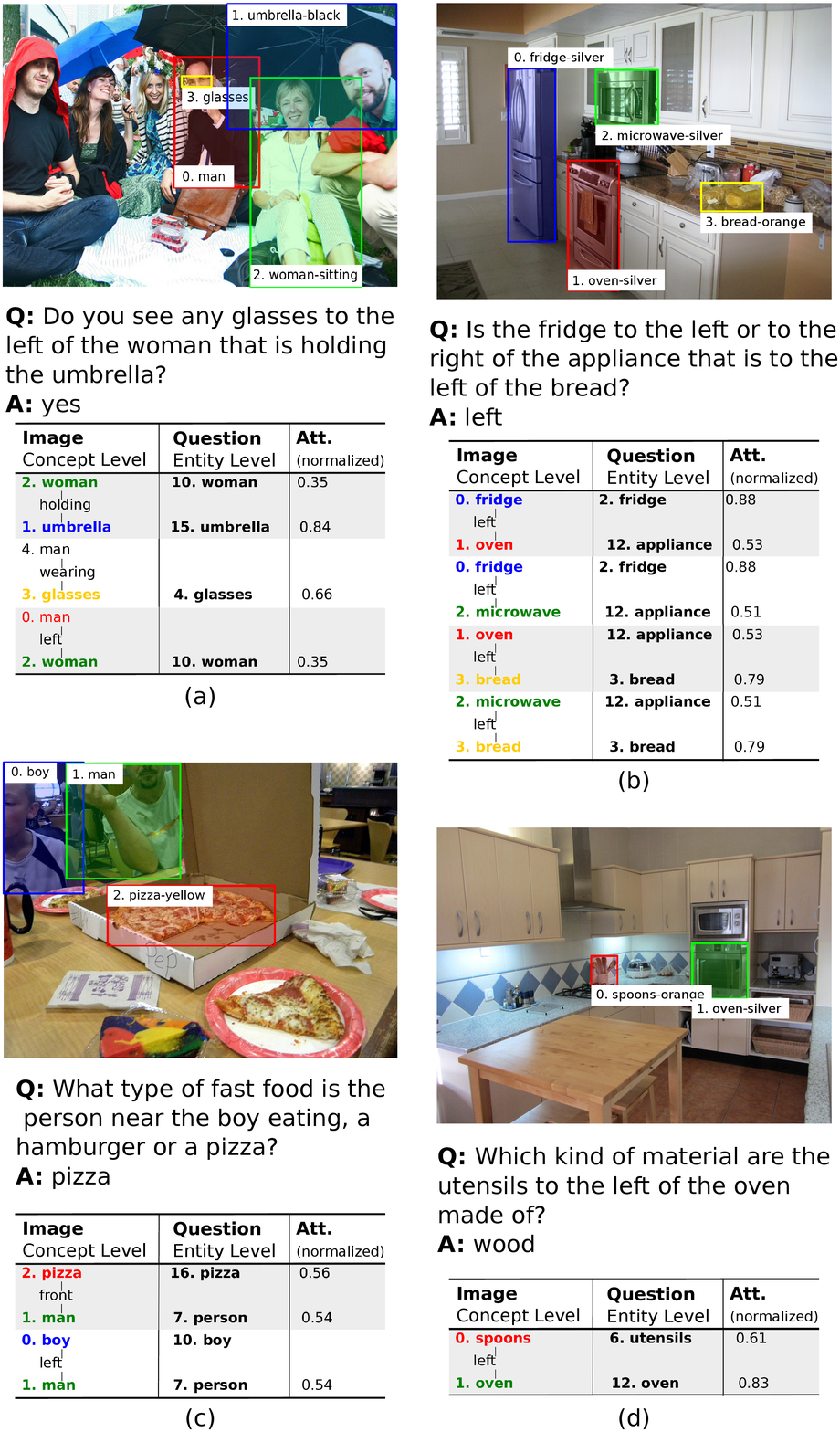}
\caption{Visualization of our MGA-VQA multi-modality alignment. The attention value is the normalized learned attention of the last layer. The bounding boxes with IDs in the images correspond to the text with the same color in the table.}
\label{fig:Transformer_attentions}
\end{figure}
%%%%%%%%%%%%%%%%%%%%%%%%%%%%%%%%%%%%%%%%%%

\subsubsection{Ablation study}
% \textbf{[Ablation study]}%\\
To validate each component of our model, we conducted experiments for an ablation study, and the results are shown in Table~\ref{table:ablation}.
The columns ``Single Modality'' and ``Multi-Modality'' represent whether features from both the image and the question are used for the final answer prediction.
The single modality setting uses only the question as the input, which tests whether the model's good performance comes from the language bias of the dataset.
Since there are no image features under the single modality setting, the multi-granularity alignment is not applicable.
For the multi-modality setting, three experiments were conducted in order to verify each level's granularity alignment.
To guarantee fair comparisons, we always set three different initialized Transformer modules for each single-level granularity alignment, preventing the model scale from affecting the experimental result.
To address the concerns about if the current merging method involves redundancy in concept-entity level, we set experiment about ``Node Reduction'' by merging the same entity nodes in two graphs into one node and combining their connecting edges.

Experiments 21, 22, and 23 show the performance of using a single alignment for multi-modality, validating the need for multiple granularity level alignments.
In addition, more abstract information leads to a higher overall accuracy, which may be due to achieving a better and easier alignment. 
However, when there is inaccurate information in the high-level granularity alignment, the lower-level granularity alignment will help provide some of the details.
More examples of this are shown in Section~\ref{section-quali}.
Experiment 24 shows that the good performance is not from the bias of the dataset.
Experiment 25 is used to test the validity of the lead graph, and the results indicate that with such guidance, the overall performance is improved.
Experiment 26 shows that the tokens in different modalities play their roles, and removing them leads to bad performance. 
To further validate our model's reasoning ability, we train and evaluate our model with the ground-truth scene graphs and achieve 92.79\% result.
Since there is no annotated scene information of the testing split, we randomly divided the validation split into two, one for validation and the other for testing.

\subsection{Quantitative results on VQA}
We also evaluate our model on VQA-v2 dataset for both test-dev and test-std splits, and the results are summarized in Table~\ref{table:vqa}
In addition to the methods tested with GQA, we also test the following methods.
Counter~\cite{zhang2018learning} aims to solve object counting problem.
MUTAN~\cite{ben2017mutan}, BLOCK~\cite{ben2019block}, MUREL~\cite{cadene2019murel} and MCAN~\cite{yu2019deep} mainly focus on different multi-modal fusions.
VL-BERT~\cite{su2019vl} and ViLBERT~\cite{lu2019vilbert} are Transformer-based models that require extra pre-training with extra, large-scale training data.

\subsection{Qualitative results}
\label{section-quali}
To better evaluate our model, we visualize some examples of our multi-modal alignment in Figure~\ref{fig:Transformer_attentions}.
The table shows the alignment between the concept-level components in the image and the entity-level components in the question. 
The corresponding object regions are highlighted in the image.
The results highlight the promising performance of our method; our model turns out to be able to find correlated elements for both modalities, and even at the image concept level, if the attributes are not accurately detected, with the assistance of the low-level alignment, the model obtains the correct final results (i.e., in Figure~\ref{fig:Transformer_attentions} (d), the attribute of the spoon is not detected as wood.)
\section{Discussion and Conclusion}
\label{section-con}
In this work, we propose a novel architecture named MGA-VQA.
In contrast to previous works that align visual and question features at a single level, our proposed Transformer-based model achieves multi-granularity alignment and jointly learns the intra- and inter-modality correlations.
In addition, we construct a decision fusion module to merge the outputs from different granularity Transformer modules.
In experiments, our model achieves outperforming results in GQA dataset and decent results in VQA-v2 dataset. 
% In experiments, our model outperforms state-of-the-art methods on the GQA dataset and non-pretraining methods on the VQA-v2 dataset. Additionally, our model achieves competitive results with pre-trained models that need extra datasets and computational resources.
Furthermore, we conduct ablation studies to quantify the role of each component in our model.

% Pre-training has become a nascent research direction in vision-language community. 
% External datasets~\cite{sharma2018conceptual,lin2014microsoft} are used to pre-train the BERT framework, and generally contain millions of images and captions.

Research on graph-based VQA remains ongoing.
% One direction is building a better multiple granularity decision fusion module.
One direction is involving pre-training mechanisms into current multiple granularity alignment and building a better decision fusion module.
The other direction is finding a strategy to better overcome misdetection or incorrect detections at a single granularity level.
Our MGA-VQA is an attempt to explore this issue. However, we expect that additional, related research will be conducted in the future.

%%%%%%%%% REFERENCES
{\small
\bibliographystyle{ieee_fullname}
\bibliography{PaperForReview}
}

\end{document}